\pdfoutput=1
\RequirePackage[T1]{fontenc}
\documentclass[letterpaper,10pt,conference]{ieeeconf}
\IEEEoverridecommandlockouts
\usepackage{times}
\usepackage{cite}
\usepackage{url}
\usepackage{amsmath}
\usepackage{graphicx}
\usepackage{booktabs,tabularx}
\usepackage{float}
\usepackage[hidelinks]{hyperref}
\hypersetup{pdftitle={Hand2Robot: An End-user Friendly Demonstration Collection Framework to Facilitate Home Deployment of Imitation Learning},
  pdfauthor={Moniruzzaman Akash, Sujosh Nag, Sajay Arthanat, Momotaz Begum}}

\title{\LARGE\bf
Hand2Robot: An End-user Friendly Demonstration Collection Framework to Facilitate Home Deployment of Imitation Learning}
\author{Moniruzzaman Akash$^{1,*}$, Sujosh Nag$^{1}$, Sajay Arthanat$^{2}$, and Momotaz Begum$^{1}$%
\thanks{$^{1}$Dept.\ of Computer Science and $^{2}$Dept.\ of Occupational Therapy,
University of New Hampshire, USA.
\texttt{\{moniruzzaman.akash, sujosh.nag, sajay.arthanat, momotaz.begum\}@unh.edu}. $^{*}$Corresponding author.}}

\begin{document}
\maketitle
\thispagestyle{empty}
\pagestyle{empty}

\begin{abstract}
We present Hand2Robot (H2R), a pipeline that lets users demonstrate tasks
with their own hands and converts egocentric RGB-D recordings into
visuomotor policy training data. The key idea is to use how and when
a person grasps an object to guide conversion to a parallel-jaw gripper.
Six grasp-specific contact frames define the gripper's pose, while predicted
manipulation phases select task-relevant motion and time grasp and release.
Replaying the trajectories in Gaussian-splat scenes supplies matching robot
images, and calibrated surface electromyography supervises how firmly to
hold objects. We evaluate seven tasks in the lab, Home Lab, and a real home;
on the six lab tasks, force-enabled policies achieve 88.9\% macro-average
success on Spot and 87.8\% on Stretch~4. In a Cube task study with three adults, H2R demonstrations
reach 86.7\% without force supervision, against 45.6\% for SpaceMouse and
12.2\% for Meta Quest~3. On unseen objects, the contact frames complete
32/36 replays versus 16/36 and 15/36 for single-rule frames adapted from
prior work. These results support natural human demonstration as a source of
useful robot training data without requiring users to operate a
teleoperation interface.
\end{abstract}

\vspace{-0.5em}
\section{Introduction}

\begin{figure*}[!t]
  \centering
  \vspace{0.5em}
  \includegraphics[width=0.8\textwidth]{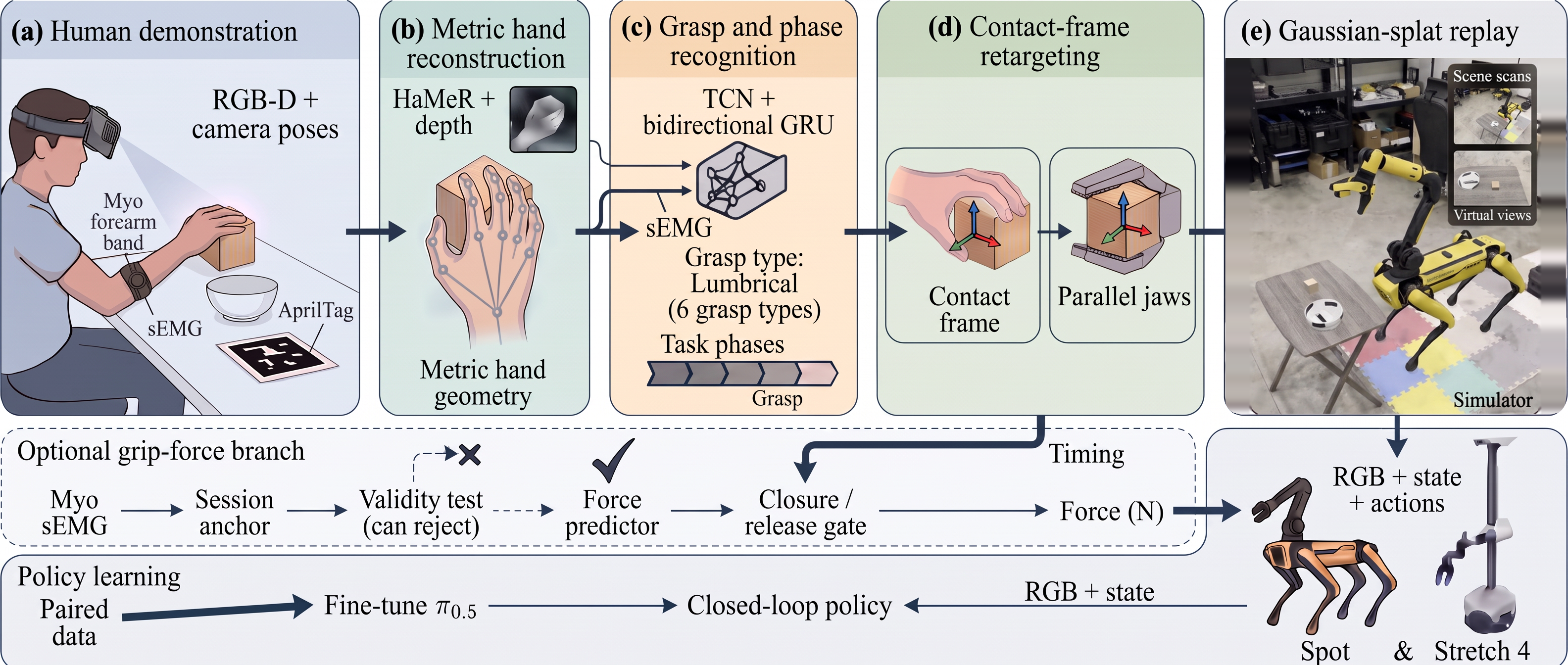}
  \caption{H2R overview: grasp- and phase-conditioned retargeting,
  Gaussian-splat replay, and optional sEMG grip-force labels convert human
  demonstrations into robot policy training data. Illustrations are schematic.}
  \label{fig:pipeline_overview}
\vspace{-0.2em}
\end{figure*}
\label{sec:introduction}
Foundational models (FMs) in imitation learning (IL) -- e.g., VLAs \cite{kim2024openvla}, $\pi$ family \cite{pi2025pi05} -- can struggle with novel setups \cite{kim2025oft}. Fine-tuning with task-specific demonstrations can improve performance in new settings \cite{kim2024openvla}. In case of home deployment of IL-enabled robots to support eldercare -- a much-discussed application domain for robots ~\cite{lee2024preferences} -- lay users, such as older adults and their family members, could provide these on-site demonstrations. In this case, data quality is a key bottleneck: demonstration quality affects policy reliability \cite{belkhale2023dataquality}, and lay-user demonstrations are typically erroneous \cite{iros2024}. Observation--action data for visuomotor policies are commonly collected by teleoperating physical robots using devices such as SpaceMouse~\cite{chi2023diffusionpolicy} or virtual-reality (VR) controllers, which can impose substantial demands on
non-expert users, with success and workload depending on the interface and
task~\cite{zhou2026static}.
This motivates Hand2Robot (H2R), which lets users demonstrate tasks without controlling a robot.

H2R collects egocentric videos of users performing tasks naturally and converts them into observation--action pairs for parallel-jaw robot policies, evaluated here on Spot and Stretch~4. It addresses the classical \textbf{\textit{correspondence problem}}: translating an observed behavior into actions that achieve corresponding effects with a different embodiment~\cite{nehaniv2001like}.

The three subproblems of the \textit{correspondence problem} involved with mapping egocentric human demonstrations to robot actions are challenges -- \textbf{action correspondence}, \textbf{perceptual correspondence}, and \textbf{embodiment mismatch}.

\textbf{Action correspondence} This is the challenge of mapping human actions onto robot actions given the difference in morphology, kinematics, and dynamics. The first challenge is to identify the human actions that need to and can be mapped  -- e.g. a human may begin a task with hands outside of the robot's workspace -- an impossible mapping for a robot, or pause before making contact -- a not-needed mapping. The next challenge is doing a high-fidelity transfer of human grasps to a parallel jaw gripper grasp. In an activities-of-daily-living study, the six grasp types used here together account for 83.3\% of estimated daily hand-use time~\cite{vergara2014grasps}. Failure to retarget the nuances in these very different grasps accurately to a parallel jaw gripper can contribute to task failure. The final challenge is generalization of this action correspondence pipeline to different robot types.

\textbf{Perceptual correspondence}: This is the challenge of converting human egocentric videos to observation-action pairs that a robot can directly use for policy training. The human hand and moving head-mounted camera produce images that differ from what a robot will observe at run time. Transferring them to the robot's perceptual space must account for these differences while preserving object motion and aligning each image with the corresponding robot action~\cite{lepert2025phantom,freeman2026warped}. Generating observations from a robot-mounted camera also requires accounting for viewpoint differences~\cite{freeman2026warped}.

\textbf{Embodiment mismatch}: This is about accounting for these differences in human and robot sensing. Vision captures motion and appearance but does not directly measure how firmly a person holds an object. Force-related sensing during demonstrations can therefore complement visual supervision, provided its measurements are mapped to the robot's gripper commands.

Using egocentric human videos for IL policy training has recently gained
traction~\cite{kareer2025egomimic,lepert2025phantom}.
Approaches to action correspondence differ in what they transfer.
Fixed thumb--index mappings simplify parallel-jaw retargeting but do not
distinguish natural grasp types~\cite{liu2025egozero,wang2026humanego,zhang2026easymimic}.
Learning from aligned human and robot data reduces the embodiment gap but
retains a need for robot demonstrations~\cite{kareer2025egomimic,zheng2026egoscale}.
Dexterous-hand retargeting preserves finger-level detail, but does not directly
specify contacts for two opposing jaws~\cite{wu2026hug,li2025okami}.
Object-centric approaches instead use estimated object poses or reconstructed
hand--object interactions to guide motion~\cite{li2025okami,wang2026egoinfinity}.
Errors in this geometry can shift the intended contact; simulation-based
refinement adds computation and still depends on reliable object-state
estimation~\cite{liu2026egoengine}.
These limitations motivate our focus on grasp-specific parallel-jaw mappings
that preserve natural interaction without requiring a tracked object trajectory.

Approaches to perceptual correspondence fall into three groups.
Image-editing methods mask, randomize, or replace the human embodiment to
reduce appearance differences~\cite{kareer2025egomimic,zhang2026easymimic,lepert2025phantom},
but these edits alone do not establish the 3D geometry of unseen surfaces.
Shared spatial representations, such as points or hand--object tokens,
reduce appearance dependence but pass localization and tracking errors
directly to the policy~\cite{liu2025egozero,wang2026humanego}.
Scene-reconstruction methods render robot observations from new viewpoints,
but errors in object poses can misplace rendered objects, particularly
under occlusion~\cite{liu2026egoengine,freeman2026warped}.
The remaining challenge is to generate observations consistent with the
mapped actions across viewpoints without relying on continuous object tracking.

Most approaches discussed above emphasize visual and geometric correspondence. ForceBand~\cite{he2026forceband} shows that surface electromyography (sEMG) can also enrich human demonstrations with force information.

H2R integrates grasp-specific action conversion, rendered robot observations,
and grip-force supervision into a pipeline for learning from natural human
demonstrations. Our contributions are:
\begin{itemize}
\item Action correspondence: Five predicted manipulation phases select task-relevant motion and grasp--release timing; six grasp-specific contact frames define parallel-jaw contact poses. A robot-specific tool transform adapts the contact trajectory without continuous object-pose tracking.
\item Perceptual correspondence: H2R pairs Gaussian-splat rendered observations~\cite{qureshi2025splatsim} with grasp- and phase-conditioned actions. Replay supplies robot-camera views without additional human demonstrations.
\item Embodiment mismatch: Calibrated eight-channel sEMG estimates human grip force, providing supervision for how firmly the robot should hold an object.
\item Evaluation on Spot and Stretch~4 covers six expert-demonstrated lab tasks, a lay-user collection-interface comparison, and tomato manipulation from older-adult demonstrations in a Home Lab and one real home.
\end{itemize}

\section{H2R: A Robot-free Demo Collection Pipeline}
\label{sec:method}

Figure~\ref{fig:pipeline_overview} summarizes H2R's conversion of egocentric
demonstrations into observation--action pairs for parallel-jaw robot policies:
five stages, (a)--(e), then the optional grip-force branch and policy
learning.
\vspace{-0.5em}
\subsection{Human Demonstration}
\label{sec:capture}
A headband-mounted iPhone leaves both hands free. Our iOS application records
egocentric RGB at $1920\times1440$, LiDAR depth at $256\times192$, camera
intrinsics, and camera poses at 30\,Hz; a stationary
AprilTag~\cite{olson2011apriltag} defines the workspace frame $G$.
Force-enabled collection adds a Myo forearm armband (Thalmic Labs MYOD5)
streaming eight-channel sEMG to the same app (nominally 200\,Hz). The participant first records two or three short calibration takes
with a Vernier Go Direct Hand Dynamometer (GDX-HD) -- the \emph{session
anchor} -- then performs the task without it.

A demonstration is $\mathcal H=\{I_t,D_t,{}^G T_{C_t},e_t\}_{t=1}^{T}$: color
and depth frames, camera pose, and aligned sEMG when available; throughout,
$ {}^A T_B$ maps coordinates in frame $B$ into frame $A$. The sEMG stream is
aligned to video by recording start and duration, and its force calibration
also estimates the EMG--force lag. The output is rendered robot observations
and actions $\mathcal R=\{(o_k,a_k)\}_{k=1}^{N-1}$, optionally with a scalar
force label. \textbf{No robot is operated during capture.}
\vspace{-0.5em}
\subsection{Metric Hand Reconstruction}
\label{sec:hand_reconstruction}

HaMeR~\cite{pavlakos2024hamer} recovers a MANO hand model
\cite{romero2017mano} from each color frame. We smooth its joint rotations on
the rotation manifold and fit metric scale and translation to depth: for
camera-frame MANO joints $\hat p_{t,j}$ and valid depth-derived points
$d_{t,j}$ at the same image locations,
$(s_t,b_t)=\arg\min_{s>0,b}\sum_{j\in\mathcal V_t}
\|s\hat p_{t,j}+b-d_{t,j}\|_2^2$, a fit that preserves camera-axis
orientation. Each joint is then transformed by
$ {}^G T_{C_t}$ into the common workspace, where bounded gaps are interpolated,
local median statistics reject outliers, and symmetric low-pass filtering
reduces jitter. The result is metric hand
geometry $P_t=\{p^G_{t,j}\}_{j=1}^{21}$ for retargeting and a smoothed
articulation sequence for recognition.
\vspace{-0.5em}
\subsection{Grasp and Phase Recognition}
\label{sec:interaction_inference}

We train a recognizer on 18.9 minutes of labeled demonstrations:
128 clips from five demonstrators spanning six grasp types and 55 objects,
each with a grasp-type label and manually finalized phase boundaries
converted into framewise supervision.
\begin{figure}[H]
  \centering
  \includegraphics[width=0.8\columnwidth]{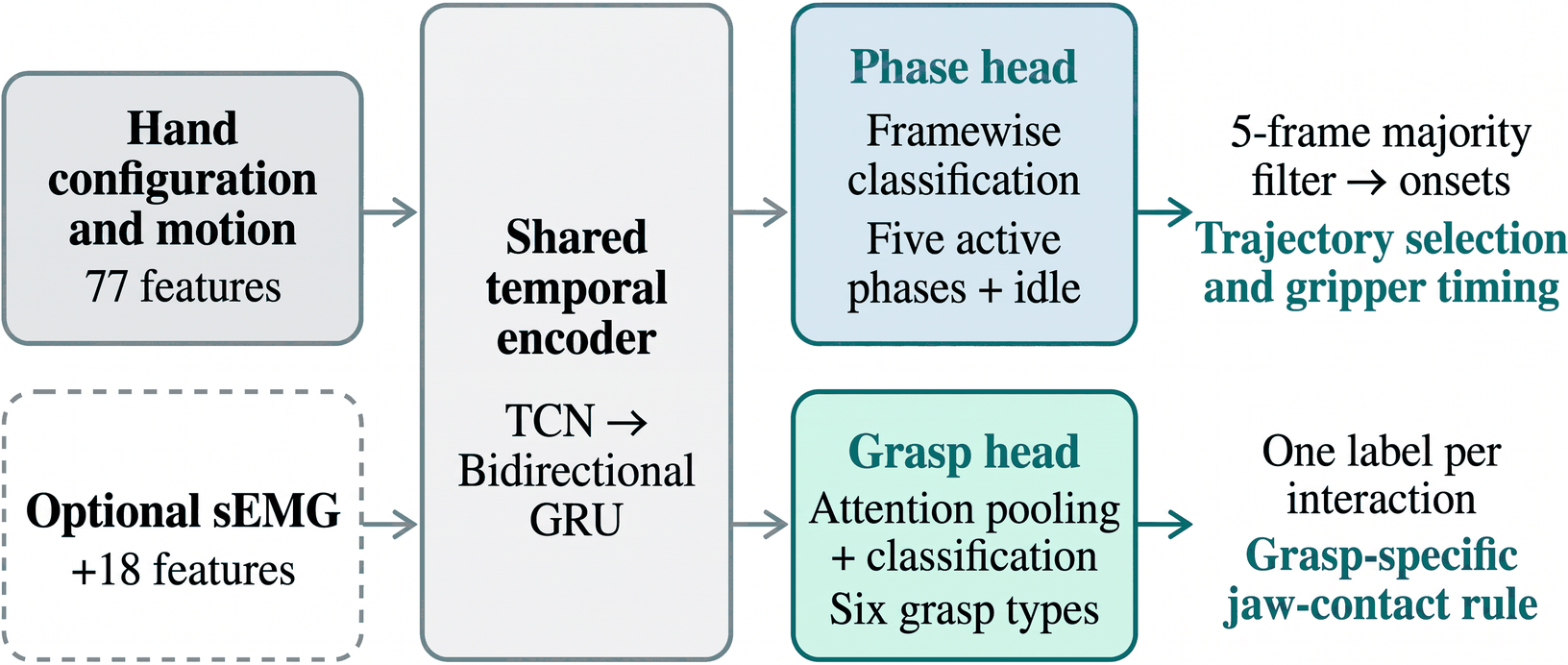}
  \caption{Offline recognition jointly predicts phase timing and grasp type
  for retargeting.}
  \label{fig:recognition_pipeline}
\vspace{-0.2em}
\end{figure}
\vspace{-0.5em}
The recognizer predicts one grasp type per interaction and one phase label per
frame (Fig.~\ref{fig:recognition_pipeline}). Its hand features exclude
absolute position, global orientation, and elapsed time, so it cannot
key on where or when an interaction happens; what remains is articulation,
joint and wrist speeds, and scale-normalized descriptors of aperture,
fingertip-to-palm distance, and finger flexion. When the armband is worn, sEMG
activation features from 150\,ms trailing windows~\cite{hudgins1993emg} are
appended and normalized per demonstration, so recognition depends on relative
activation, unlike force estimation, which preserves calibrated amplitude
(Sec.~\ref{sec:force_method}). A shared temporal encoder feeds a framewise head
over five active phases -- \emph{approach}, \emph{commit} (final alignment
before closure), \emph{grasp}, \emph{transport/task}, and \emph{release} -- plus an \emph{idle}
label for motion outside the interaction, and an attention-pooled head that
assigns one of the six grasp types in Fig.~\ref{fig:retargeting_overlay}. Its
bidirectional context makes recognition an offline operation. The heads train
under a single objective,
$\mathcal L_{\mathrm{rec}}=\sum_t w_{y_t^p}\,\mathrm{CE}(\hat y_t^p,y_t^p)
+\mathrm{CE}(\hat y^g,y^g)$, over the per-frame phase labels $y_t^p$ and the
interaction label $y^g$. The inverse-frequency weight
$w_{y_t^p}$ stops long transport/task intervals from dominating the brief grasp
transitions that set gripper timing. Training adds temporal resampling by
factors of 0.7--1.4 and Gaussian feature jitter; the final models are fitted on
all 128 annotated demonstrations.

At inference, a five-frame majority filter suppresses isolated phase changes,
and the first occurrence of each active phase within an interaction supplies
its onset; an absent phase yields no boundary. The grasp type selects the
geometric rule, while commit, grasp, and release onsets set trajectory
selection and gripper timing (Sec.~\ref{sec:phase_trajectory}). For
multi-object tasks, the predicted phases and grasps identify successive
interactions, each with its own grasp type and boundaries.

\subsection{Contact-Frame Retargeting}
\label{sec:retargeting}

\begin{figure}[t]
  \centering
  \vspace{0.5em}
  \includegraphics[width=\columnwidth]{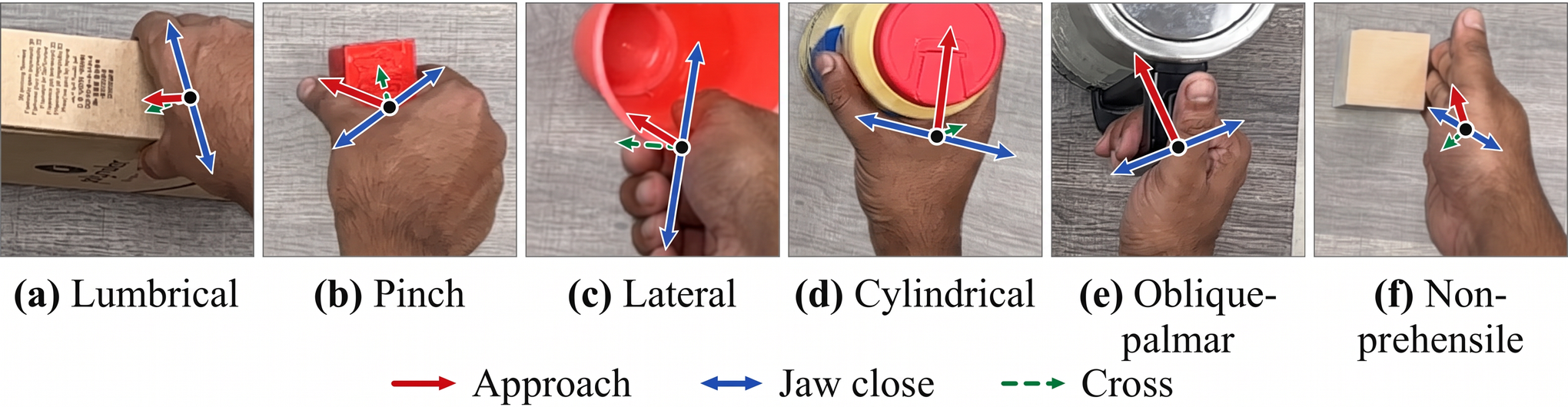}
  \vspace{-0.9em}
  \caption{H2R's six grasp types, each shown with the jaw-contact frame. The approach axis, the jaw-closing axis and
  their cross product are drawn at the contact origin and projected into the demonstration frame.}
  \label{fig:retargeting_overlay}
\vspace{-0.5em}
\end{figure}

Each rule reads the metric MANO joints $P_t$ and returns a contact origin
$c_t^g$, where the jaws should close, plus two direction cues $u_t^g,v_t^g$
between named joints. A shared operator then normalizes the first cue, projects
the second perpendicular to it, and crosses them for the third; the grasp type
$g$ decides which cue becomes the approach axis and which the jaw-closing axis:
\begin{equation}
{}^G T_{J_t}=
\begin{bmatrix}
\operatorname{Frame}(u_t^g,v_t^g) & c_t^g\\
0 & 1
\end{bmatrix}.
\label{eq:retarget}
\end{equation}
$J$ is a robot-independent jaw-contact frame, its axes ordered approach, jaw
closing, cross product. Signs follow the previous frame, so a geometrically
equivalent $180^\circ$ flip never becomes a discontinuous robot command, and
missing landmarks fall back to the same finger segment or the last valid
orientation.

The six rules differ only in which landmarks supply the origin and the cues
(Fig.~\ref{fig:retargeting_overlay}). In the precision grasps --
\textbf{lumbrical}, \textbf{pinch}, and \textbf{lateral} -- the origin lies at
the thumb--finger contact, and the opposing finger directions and the span
across the finger bases supply the cues. The power grasps,
\textbf{cylindrical} and \textbf{oblique-palmar}, take the origin at the palm,
shifted toward the wrapped fingers for a handle, and orient from the palm and
finger-wrap directions. \textbf{Non-prehensile} uses the palm face as a pushing
surface and holds the gripper in a fixed configuration. The rules run on the
30\,Hz hand sequence before robot resampling; target-robot reach and jaw width
are checked later.

\subsubsection{Phase-aware trajectory and robot adaptation}
\label{sec:phase_trajectory}

We resample the jaw-contact trajectory to a 10\,Hz control grid. With
$i_m,i_c,i_r$ the commit, grasp, and release onset indices of an interaction,
we discard motion before commit and the final half of release, removing
unconstrained entry and withdrawal. A new approach connects the robot's
starting pose $(p_0,q_0)$ to the retargeted commit pose $(p_m,q_m)$, both
jaw-contact poses in $G$. With $\Delta t=0.1$\,s and translation speed
$v=0.25$\,m/s it spans $n_a=\max(1,\lceil\|p_m-p_0\|/(v\Delta t)\rceil)$ steps
that interpolate position linearly, $p_k=(1-\alpha_k)p_0+\alpha_kp_m$, and
orientation by $\operatorname{Slerp}(q_0,q_m,\alpha_k)$, with $\alpha_k=k/n_a$.
We write the resulting jaw-contact sequence $ {}^G\widetilde T_{J_k}$ and
reindex its phase boundaries on the new timeline; later contact and task motion
keeps its relative timing. In multi-interaction tasks the first approach
starts at home, and later approaches start from the previous segment's
terminal pose.

For prehensile interactions the jaw command follows open--closed--open, with
closure starting at $i_c$. Release is refined by the strongest opening edge in
normalized hand aperture, smoothed over five frames, within 0.60\,s of $i_r$
and after closure; if the maximum opening rate is below
$0.25\,\mathrm{s}^{-1}$, the phase boundary is retained. For downward
placement, release is delayed while the end-effector continues descending, for
at most a further 0.60\,s. The resulting command is $\gamma_k=0$ for
$i_c\leq k<i_o$ and $\gamma_k=1$ otherwise,
where $i_o$ is the refined release index and 0/1 denote closed/open, so finger
adjustments during transport create no extra releases. Missing grasp or release
boundaries leave the gripper open and flag the segment for review. The rule is
applied separately to each prehensile interaction.

For robot tool frame $E$ and base frame $B$, adaptation uses
$ {}^B T_{E_k}=({}^G T_B)^{-1}\,{}^G\widetilde T_{J_k}\,{}^J T_E$.
The fixed $ {}^J T_E$ accounts for wrist-to-contact offset and each gripper's
axis conventions. The base placement $ {}^G T_B$ is specified during simulation
and measured against the workspace fiducial at deployment (Spot's docking-station tag); for Spot it can use
the inertial vision frame $V$, $ {}^G T_B={}^G T_V{}^V T_B$. The transformation preserves
timestamps and gripper commands. Robot-specific inverse kinematics (IK) checks
reachability and joint limits, interpolating isolated failed poses between
earlier and later valid IK solutions; demonstrations with IK failures in more than
20\% of retained frames are discarded.
\vspace{-0.5em}
\subsection{Gaussian-Splat Replay}
\label{sec:replay}

Our simulator separates appearance from contact: Gaussian-splat scans of the
workspace, robot, and objects render~\cite{kerbl20233dgs,qureshi2025splatsim},
while robot kinematics and simplified collision geometry carry motion and
rigid-body interaction, with only rigid transforms crossing between them. Fixed
scene and AprilTag transforms place the scanned assets, and robot alignment
uses the measured base placement $ {}^G T_B$ rather than visual localization in
the splat. The retargeted trajectory then drives the robot's inverse kinematics
and transforms its link splats at each step, while a held object follows the
hand kinematically rather than through simulated contact: the grasp event fixes
a relative transform $ {}^J T_O$, so
$ {}^G T_{O_k}={}^G\widetilde T_{J_k}\,{}^J T_O$ for $i_c\leq k<i_o$. Its
endpoints come from the retargeted contact events and asset geometry, and
release seats the object on the surface that supports it there, such as the
interior of a bowl rather than the table beneath. Each interaction uses its own
assets and grasp--release events, so replay needs scene preparation and object
geometry but no per-frame object tracking. Liquid and salt particles are
omitted; container motion is preserved.

External and wrist-mounted cameras render RGB observations, the wrist view set
by the robot's calibrated camera-to-tool transform. We sample external views
around a nominal deployment camera pose inside the reconstructed scene and vary
scene lighting, inspired by RoboSplat~\cite{yang2025robosplat}; these preserve
action labels and are intended to tolerate small changes in rollout-camera
placement and illumination, though appearance edits do not reproduce exact
physical relighting. Each step records RGB, end-effector pose, joint state,
gripper state, and timestamp, and with the base-frame tool pose $(p_k,R_k)$ the
pose-action convention is
$a_k=[p_{k+1}-p_k,\ r_6(R_k^\top R_{k+1}),\ \gamma_{k+1}]$, where $r_6$ is the
first two columns of the relative rotation and translation is in base
coordinates. Grip-force supervision appends an absolute force
target, raising the action to 11 dimensions, and the previous command supplies
the matching state channel at rollout. Export checks length consistency and a fixed object-to-gripper transform
during transport; a grasp that fails to attach aborts generation rather than
exporting an empty-gripper demonstration.
\vspace{-0.5em}
\subsection{Optional Grip-Force Branch}
\label{sec:force_method}

\paragraph{Measurement and calibration.}
sEMG adds scalar grip-force estimates to human demonstrations without
instrumenting the fingertips~\cite{bazina2025koopman,he2026forceband}. Isometric calibration
emphasizes 10--40\,N while manipulation grips peak at roughly 5--9\,N, so the
model learns where the signal is strong and operates where it is weak. Short
dynamometer recordings align the sEMG and force signals and estimate a
session-specific gain. We extract smooth activation envelopes, subtract a
per-demonstration resting baseline, and normalize by this gain, accounting for
recording conditions while preserving force-related amplitude differences.

\paragraph{Anchor validity.}
The collection recipe combines at least 50\,s of gentle squeezing in the
2--8\,N band with at least 10\,s above 15\,N, without repositioning the
armband. These coverage targets characterize the accepted anchors in our study;
coverage alone does not establish validity. Before predicting force we check
that the anchor gives consistent sEMG--force correlation, low-force coverage,
and enough activation range to estimate gain. An accepted anchor supplies the
session calibration, and a population gain provides the reference otherwise.
Section~\ref{sec:force_results} evaluates this procedure.

\paragraph{Temporal force predictor.}
A diffusion-style predictor learns from a population dataset of 138 isometric
sEMG--force recordings from nine people, paired with dynamometer measurements. Rather than regress force
directly, it learns to remove noise: training corrupts a recorded 2.56\,s force
trace and asks a dilated convolutional network, conditioned on the calibrated
sEMG for that window, for the step back to the clean trace. At inference it
starts from noise, repeats that step 24 times using sEMG alone, and averages
eight samples. The predicted force $\hat F$ is aligned to the 10\,Hz trajectory and retained
only within each prehensile interaction's closed interval,
$f_k=\mathbf{1}[i_c\leq k<i_o]\,\hat F_k$, so approach and non-prehensile
segments receive zero grip-force labels. The scalar is expressed in
dynamometer-referenced newtons, not contact pressure.

\subsection{Policy Learning}
\label{sec:policy_learning}

Both Spot and Stretch~4 expose gripper joint effort rather than force in
newtons. For each robot we measure force across effort commands with the same
Vernier dynamometer, using a 3D-printed cap that accommodates gripper contact;
the measured effort--force mapping converts the policy's force target to an
effort command without an assumed conversion factor, though this calibration
does not provide online force or tactile feedback. We fine-tune
$\pi_{0.5}$~\cite{pi2025pi05} on rendered images, robot state, task language,
and the aligned action targets, with a robot-specific data adapter converting
the exported pose representation to the policy's action convention. At
deployment the policy receives robot camera images and proprioception and
executes predicted action chunks in closed loop, while hand reconstruction,
phase recognition, and scene rendering remain offline data-generation
operations.

\section{Experiments}
\label{sec:experiments}

We ask five questions. \textbf{Q1:} do policies trained from converted
demonstrations execute multi-step tasks on more than one robot?
\textbf{Q2:} for lay users, does H2R produce better policy data than
controller-based collection? \textbf{Q3:} does grip-force supervision improve
task success? \textbf{Q4:} do grasp-specific contact frames execute better
than single-rule frames on unseen objects? \textbf{Q5:} are the remaining
components accurate enough to support those results? Evaluation covers seven tasks: six lab tasks and tomato
pick-and-place, with the interface comparison reusing Cube.
\vspace{-0.5em}

\subsection{Evaluation Protocol}

We use Boston Dynamics Spot and Hello Robot Stretch~4 with parallel-jaw
grippers. Each H2R training dataset contains 20 original demonstrations
converted to robot observations and actions, plus 80 augmented versions with
changes to the external camera pose and scene lighting, for 100 demonstrations
in total. Augmented versions reuse the source trajectories rather than
additional human recordings. We fine-tune $\pi_{0.5}$ for 30,000 steps with
batch size 8, using matched training settings within each task and robot.
Force-off and force-on conditions use the same
source demonstrations, retargeting, and rendered observations, differing
only in grip-force supervision. Calibration anchors are excluded from
task-demonstration counts.

Each policy condition in Table~\ref{tab:lab} and
Fig.~\ref{fig:older_adult_policy_results} is evaluated
over \textbf{30 physical rollouts}.
Success requires completing all task steps in order without manual
intervention; partial completion does not count. Averages use rollout
counts before rounding and weight tasks or participants equally.
Differences are reported in percentage points. Thirty rollouts give a
resolution of 3.33 percentage points per condition, so differences of one or
two successes are not established improvements. Repeated rollouts
characterize one trained policy; they are neither independent training runs
nor additional participants, so we keep results at the participant level and
read the comparisons descriptively.
\vspace{-0.5em}
\subsection{Expert Demonstrations: Lab Tasks}

\begin{figure}[t]
\centering
\vspace{0.2em}
\includegraphics[width=\columnwidth]{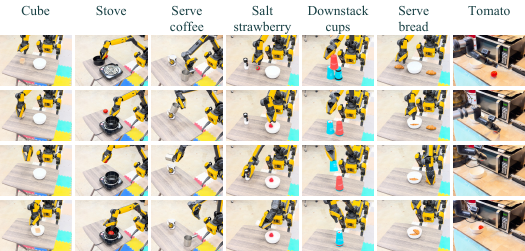}
\caption{Rollout keyframes. Each column is one task and time progresses top to
bottom: six lab tasks and real-home tomato manipulation.}
\label{fig:task_rollouts}
\vspace{-0.2em}
\end{figure}

Addressing Q1 and Q3, expert users collected the six lab-task datasets in
Table~\ref{tab:lab}, with rollout keyframes in Fig.~\ref{fig:task_rollouts}.
The tasks cover five prehensile grasp types and non-prehensile pushing:
\textbf{Cube:} pick up a wooden cube with a lumbrical grasp and release it
into a bowl.
\textbf{Stove:} pick up a pan with a lateral grasp and place it on the
stove, then pick up a tomato with a lumbrical grasp and release it into the pan.
\textbf{Serve coffee:} pick up a coffee mixer jar with an oblique-palmar
grasp and pour into a cup, then push the filled cup toward the recipient with
a non-prehensile push.
\textbf{Salt strawberry:} place a strawberry in a bowl with a pinch grasp,
then pick up a salt jar with a cylindrical grasp and dispense salt into it.
\textbf{Downstack cups:} push a stack of three cups apart non-prehensilely,
then lift the remaining cup with a lumbrical grasp onto the others.
\textbf{Serve bread:} transfer two croissants from the table to a bowl.

Table~\ref{tab:lab} reports complete task execution, including transitions
between grasp types and between grasping and pushing. Policies trained from
converted demonstrations succeed on all six tasks on both robots, so the
retargeted trajectories and rendered observations carry enough information
for multi-step interaction. Grip-force supervision raises the macro-average
on both robots and helps most on Stove, which nonetheless remains the
lowest-success task; several tasks are unchanged and one decreases. The
gain is therefore aggregate rather than uniform.

\begin{table}[t]
\centering
\caption{Lab-task success rates (\%) ($n=30$).}
\label{tab:lab}
\small
\setlength{\tabcolsep}{3pt}
\renewcommand{\arraystretch}{1.12}
\begin{tabularx}{\columnwidth}{@{}Xcccc@{}}
\toprule
& \multicolumn{2}{c}{Spot} & \multicolumn{2}{c}{Stretch~4}\\
\cmidrule(lr){2-3}\cmidrule(l){4-5}
Task & Force off & Force on & Force off & Force on\\
\midrule
Cube & \textbf{96.67} & \textbf{96.67} & 86.67 & \textbf{93.33}\\
Stove & 53.33 & \textbf{73.33} & 46.67 & \textbf{70.00}\\
Serve coffee & \textbf{86.67} & 83.33 & 76.67 & \textbf{80.00}\\
Salt strawberry & 86.67 & \textbf{90.00} & \textbf{90.00} & \textbf{90.00}\\
Downstack cups & \textbf{90.00} & \textbf{90.00} & 86.67 & \textbf{93.33}\\
Serve bread & 96.67 & \textbf{100.00} & 93.33 & \textbf{100.00}\\
\midrule
Macro-average & 85.00 & \textbf{88.89} & 80.00 & \textbf{87.78}\\
\bottomrule
\end{tabularx}
\vspace{-0.2em}
\end{table}

\vspace{-0.5em}
\subsection{Demonstration By Lay User}
\label{sec:older_adult_studies}
\begin{figure}[H]
  \centering
  \vspace{-0.5em}
  \includegraphics[width=0.70\columnwidth]{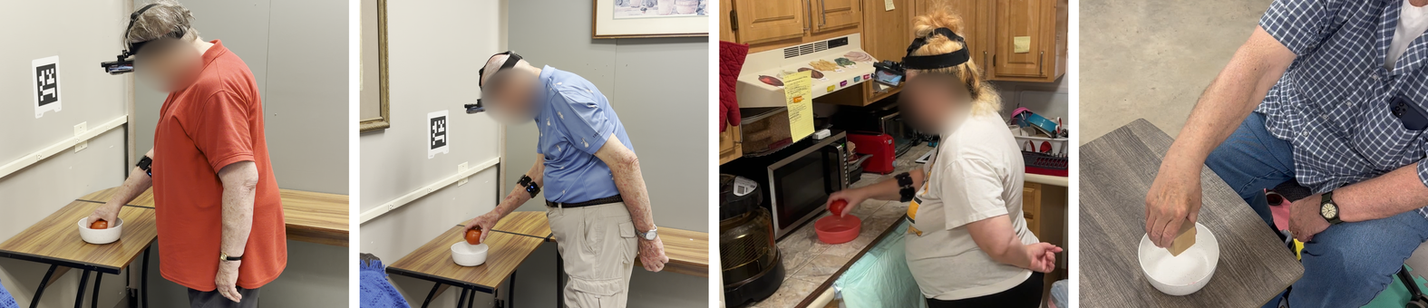}
  \vspace{-0.5em}
  \caption{Older adults demonstrating tasks with H2R.}
  \label{fig:older_adult_demonstrations}
\vspace{-0.5em}
\end{figure}

Two IRB-approved studies address Q2 and Q3 with people who are not
roboticists. Participants and datasets are separate from the expert lab
tasks. Figure~\ref{fig:older_adult_demonstrations} illustrates natural
demonstration collection.

\paragraph{Collection-interface comparison}
Three adults (C1--C3, ages 28--85 years) each demonstrate Cube with
SpaceMouse, Meta Quest~3, and H2R. None had used these teleoperation devices
before; each practised for 15 minutes, then collected 20 demonstrations per
interface. Every participant used the interfaces in the order listed above,
with at least 30 minutes between sessions; the order was not counterbalanced,
so accumulated task practice favors H2R.
Training one policy per participant--interface dataset gives nine
force-free conditions, evaluated on Spot. The controllers record robot actions
directly while H2R converts hand motion and renders observations, so the
comparison measures downstream policy quality for whole collection pipelines,
not interface usability.

\begin{figure}[t]
\centering
\includegraphics[width=0.80\columnwidth]{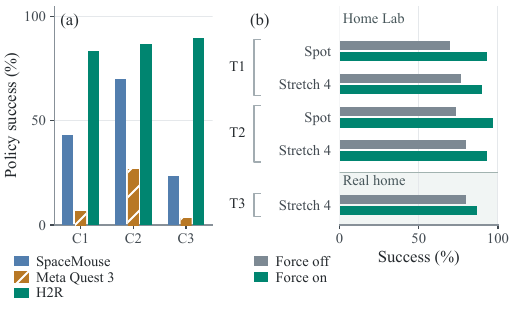}
\vspace{-0.5em}
\caption{Policy success from lay-user demonstrations ($n=30$ per condition).
(a) Cube on Spot in the lab by collection interface, without force supervision.
(b) H2R tomato manipulation with force off/on, grouped by demonstration site.
C1--C3 and T1--T3 are separate participant groups.}
\label{fig:older_adult_policy_results}
\vspace{-0.2em}
\end{figure}

H2R gives the highest success for every participant, and its three results
cluster closely while both controller conditions scatter
(Fig.~\ref{fig:older_adult_policy_results}(a)); mean success is 86.7\%
against 45.6\% for SpaceMouse and 12.2\% for Meta Quest~3. Data quality from an
unfamiliar controller thus depends heavily on the individual operator;
performing the task by hand does not. This supports moving embodiment
conversion into the data pipeline rather than asking a lay user to learn to
drive a robot.

\paragraph{Home Lab and real-home demonstrations}
A separate group (T1--T3, ages 55--87 years) demonstrates tomato
pick-and-place with H2R: T1 and T2 in the Home Lab, T3 in a real household
kitchen. No grasp type or initial hand pose is prescribed, and this study has
no controller baselines. Each participant first records the short dynamometer
calibration sequence. T1 and T2 supply Spot and Stretch~4 data, T3 Stretch~4
only, giving ten policy conditions over five participant--robot pairs.

Grip-force supervision improves success in every participant--robot
comparison, in the Home Lab and in the real home
(Fig.~\ref{fig:older_adult_policy_results}(b)). The force channel carries
information the pose and image streams leave ambiguous, and that benefit
survives the move out of a controlled setting. The real-home gain is the
smallest of the five and rests on one participant, so we read it as
directional.

\subsection{Execution on Unseen Objects}
\label{sec:unseen_objects}
\begin{figure}[t]
  \centering
  \includegraphics[width=\columnwidth]{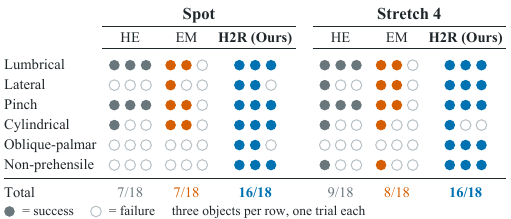}
  \caption{Unseen-object replay with only the contact-frame rule changed.
  Each token is one object, three per row, one trial each. HE is a
  HumanEgo-style single-rule frame, EM an EasyMimic-style single-rule frame,
  and H2R our grasp-specific frame.}
  \label{fig:retargeting}
\end{figure}

This experiment addresses Q4, isolating the contact-frame rule that is this
paper's central technical claim. For each grasp type, we replay the
retargeted trajectory from the same
human demonstration on three different objects absent from the
grasp-and-phase recognizer's training set. We compare H2R with single-rule
contact frames adapted from HumanEgo (HE)~\cite{wang2026humanego} and
EasyMimic (EM)~\cite{zhang2026easymimic}, holding reconstruction,
grasp-and-phase prediction, jaw timing, and robot execution settings fixed.
Only the contact-position and orientation rule changes; this tests
retargeting, not the complete baseline systems or learned-policy rollouts.
With one trial per object, robot, and method, H2R succeeds in 32/36
executions versus 16/36 for HE and 15/36 for EM, and the margin holds with
pushing excluded (26/30, 15/30, and 14/30). The gain comes from the
interactions a single pinch-like frame cannot represent, chiefly lateral and
oblique-palmar, while pinch and lumbrical are already matched
(Fig.~\ref{fig:retargeting}). Stretch~4's cylindrical failures were
attributed to its maximum gripper opening; they remain in the totals.

\subsection{Component Evaluation}

Turning to Q5, we evaluate grip-force estimation and grasp and phase
recognition, separating component accuracy from downstream policy success.

\subsubsection{Grip-force estimation from sEMG}
\label{sec:force_results}

\paragraph{Protocol.}
The corpus contains 138 isometric calibration takes (17 sessions,
84 minutes) and six dynamic sessions (S1--S6) of three anchor takes plus 129
pick-and-place demonstrations (24 minutes). The dynamometer is the manipulated
object, so force is measured during motion; the predictor receives sEMG only.
Each dynamic session is
held out in turn, with models trained on the isometric recordings and the
remaining dynamic sessions, excluding the held-out session's anchor. Results
average four seeds, five for timing. A demonstration's peak is its
98th-percentile force, peak ratio is the within-session mean of predicted over
measured peaks, and reported means weight sessions equally.

\begin{figure}[t]
  \centering
  \includegraphics[width=\columnwidth]{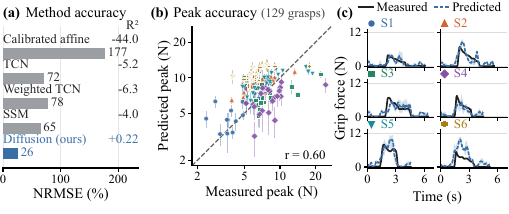}
  \caption{Population-calibrated grip-force estimation, with S1--S6 held out in
  turn. (a) NRMSE and $R^2$ per method, averaged over seeds, then
  demonstrations. (b) Diffusion peaks for all 129 demonstrations (four-seed mean
  $\pm$ SD, log--log axes); dashed line: equality, $r$: median within-session
  Pearson correlation. (c) Four-seed mean prediction (shading: seed range) for
  the demonstration nearest each session's median absolute peak error. Markers
  identify sessions in both (b) and (c).}
  \label{fig:force_results}
\vspace{-0.2em}
\end{figure}

\paragraph{Force magnitude.}
Under population calibration, diffusion reaches the lowest normalized
root-mean-square error (NRMSE) by a wide margin and is the only method with
positive $R^2$ against each demonstration's
mean force (Fig.~\ref{fig:force_results}(a)). Peak magnitude separates them
further: the calibrated affine baseline has a mean peak ratio of 5.16 against
1.15 for diffusion, and that residual overestimation means
predicted labels are not guaranteed safe robot
commands (Fig.~\ref{fig:force_results}(b)). Frame mean absolute
error (MAE) cannot rank the methods, since six differ by less than the
${\approx}1.0$\,N minimum detectable difference. Session peak ratios span
0.79--1.61 with mean frame MAE 1.40\,N, and the traces reproduce grasp onset,
duration, and release (Fig.~\ref{fig:force_results}(c)).

\paragraph{A usable calibration recipe.}
For the anchor check of Sec.~\ref{sec:force_method}, accepted anchors improve
frame MAE in 21/21 comparisons across seven methods and rejected anchors in
none of 21---42 method--session comparisons, not independent
sessions. Applying the
check tightens the values above to mean frame MAE 1.19\,N and session peak ratios of 0.77--1.28
(mean 1.05), though improved frame MAE does not imply improved peak magnitude
in every session.

\paragraph{Timing and scope.}
Median absolute grasp-onset error is 0.14\,s, with 95\% of demonstrations
within 0.5\,s. Predicted force at rest averages 0.71\,N and 7.1\% of rest
frames exceed 2\,N; the phase-locked interval in Sec.~\ref{sec:force_method}
should mask this residual activation, but its effect has not been measured end
to end. These cross-session, free-manipulation results are not comparable with
isometric, within-session, individually calibrated evaluations, nor with prior
work on other protocols~\cite{bazina2025koopman,he2026forceband}; we claim no
accuracy advantage and evaluate the force channel by downstream policy
success.

\subsubsection{Grasp and phase recognition}
\label{sec:recognition_results}

\paragraph{Protocol}
We evaluate the offline recognizer with stratified, demonstration-grouped
five-fold cross-validation on the 128 demonstrations in
Sec.~\ref{sec:interaction_inference}, averaging four model seeds. The corpus
holds 19--24 demonstrations per grasp type, captured at three tag positions
across two days, and grasp types are not tied to object identity: bowls and
pan lids each appear with several different interactions.

\begin{figure}[t]
  \centering
  \includegraphics[width=0.8\columnwidth]{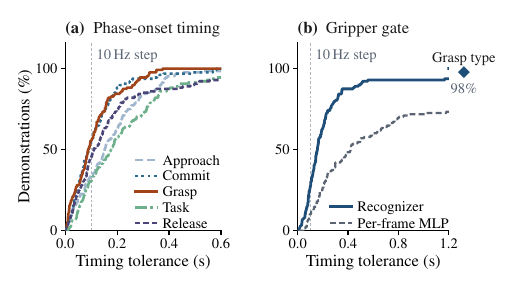}
  \vspace{-0.5em}
  \caption{Held-out recognition (128 demonstrations, four seeds).
  (a) Phase onsets within tolerance. (b) Grasp and release jointly within
  tolerance versus a per-frame MLP with identical inputs. Dashed rules mark one 10\,Hz replay step;
  the diamond denotes grasp-type accuracy.}
  \label{fig:grasp_phase_results}
\vspace{-0.2em}
\end{figure}

\paragraph{Timing the gripper schedule}
At a 0.3\,s tolerance the recognizer recovers both grasp and release onsets
on most held-out demonstrations, and grasp alone on nearly all, well above a
per-frame MLP with identical 95-dimensional inputs, folds, augmentation, and
seeds (Fig.~\ref{fig:grasp_phase_results}). Temporal context is what buys
this: it reduces mean onset error from 0.397 to 0.191\,s and raises phase
frame accuracy from 0.787 to 0.865. Grasp-onset MAE is 0.110\,s, about one
step of the 10\,Hz replay trajectory, so closure lands within the resolution
at which it is replayed. The residual error is not uniform: task onset is
consistently late and release has the largest onset MAE (0.318\,s), so
accurate closure alone does not ensure an accurate complete schedule. Timing
scores exclude never-predicted phases, at most 0.6\% of onsets for the
recognizer and 4.1\% for the MLP.

\paragraph{Grasp classification and input choice}
Grasp-type accuracy is 0.979 (Wilson 95\% interval $[0.936,0.993]$), but
classification is already saturated at this corpus scale: a
dynamic-time-warping 1-nearest-neighbor baseline reaches 0.992 without model
training. Temporal context therefore helps phase timing, not grasp
discrimination. Kinematics alone matches the full input set (grasp type 0.982,
onset MAE 0.204\,s), so recognition does not require sEMG.

\paragraph{Session shift and untested scope}
Under held-out capture-group evaluation---groups differ in both tag position
and day---phase accuracy decreases from 0.865 to 0.808 but stays above the
per-frame MLP's 0.721. All corpus demonstrations are successful and approach
from above, and demonstrator identities were not retained per clip, so
failures, other approach directions, and generalization to unseen people or
objects remain untested.

\vspace{-0.2em}
\section{Discussion and Limitations}
\label{sec:discussion}

The results support natural demonstration as a practical source of policy
training data in the evaluated settings.

The two-robot results support the combined design of grasp-specific contact
frames, phase selection, and rendered observation, but do not isolate the
contribution of individual grasp rules or of phase selection.
Rendered-to-real transfer likewise supports the observation pipeline without
establishing viewpoint invariance or superiority over other rendering
methods, and view and appearance augmentation expands the training
distribution without an established generalization benefit, which would
require held-out evaluation.

The Cube comparison suggests that familiar human actions can supervise more
effectively than controller-mediated demonstrations for these participants.
This concerns resulting policy quality rather than usability, since the
interfaces change both action and image sources. Three participants and one
task do not establish population-level accessibility, and a single-grasp task
cannot isolate the benefit of supporting six grasp types.
Turning to the force channel, the aggregate lab-task gains and the consistent
direction of the tomato comparisons support the usefulness of grip-force
labels rather than a universal benefit, and Stove's remaining difficulty shows that the channel
cannot compensate for errors in placement, sequencing, perception, or
control. Policy success does not itself establish force-estimation or
force-tracking accuracy. \textbf{Limitations.}
Parallel jaws cannot reproduce distributed finger contacts, in-hand
dexterity, or every object size. H2R relies on hand reconstruction, depth
association, phase recognition, and workspace calibration, each a source of
trajectory error, and hand occlusion remains a failure mode. Replay further
requires scanned assets, placement, and approximate contact geometry, and a
held object can follow its kinematic attachment where real friction would not
support the grasp. The current 3DGS replay covers rigid, non-articulated
assets only: deformable objects, material flow, and articulated mechanisms are
unsupported, making simulation-ready articulated reconstruction an important
direction for future work~\cite{xia2025drawer,cao2025physxanything}. The sEMG
channel is an indirect scalar estimate, not tactile feedback: calibration
addresses session gain and baseline correction resting activation, but neither
recovers contact pressure or slip. Finally, the tomato demonstrations come from
two participants in the Home Lab and one in a real home, where participant,
site, and dataset vary together, so the lower force-enabled result there cannot
be attributed to the environment alone. Broader deployment requires more
participants, homes, tasks, and independent training runs.

\vspace{-1.0em}
\section{Conclusion}
\label{sec:conclusion}

Robot training data usually comes from operating a robot. H2R removes that
requirement by moving embodiment conversion into the data pipeline: a person
performs the task by hand, and the recording becomes parallel-jaw
observations and actions via grasp-specific contact frames, phase-aware
trajectories, Gaussian-splat replay, and an optional sEMG grip-force channel.
Over the six lab tasks of a seven-task evaluation spanning the lab, Home Lab,
and a real home, force-enabled policies reach macro-averages of 88.89\% on Spot
and 87.78\% on Stretch~4; with the pipeline otherwise fixed, grasp-specific
contact frames complete more
unseen-object replays than single-rule frames from prior work; and H2R
demonstrations outperform both teleoperation interfaces in the Cube study.
The scope is three participants per study, one real home, and rigid objects;
widening it and supporting articulated and deformable assets at replay are
the next steps.

\bibliographystyle{IEEEtran}
\bibliography{references}

\end{document}